%% file: main.tex
\documentclass{article}
\usepackage{iclr2027_conference,times}
\usepackage[T1]{fontenc}
\usepackage[utf8]{inputenc}
\usepackage{microtype}
\usepackage{amsmath,amssymb,mathtools}
\usepackage{booktabs,array,tabularx}
\usepackage{enumitem}
\usepackage{graphicx,xcolor}
\usepackage{tikz}
\usetikzlibrary{arrows.meta,positioning,fit,backgrounds}
\usepackage{listings}
\usepackage[most]{tcolorbox}
\usepackage{hyperref}
\definecolor{vblue}{HTML}{2855A6}
\definecolor{vgreen}{HTML}{277153}
\definecolor{vorange}{HTML}{A76520}
\definecolor{vgray}{HTML}{F3F4F6}
\hypersetup{colorlinks=true,citecolor=vblue,linkcolor=vblue,urlcolor=vblue}
\newcolumntype{Y}{>{\raggedright\arraybackslash}X}
\newcommand{\method}{\textsc{SO-RSI}}
\newtcolorbox{researchcard}[1]{colback=vgray,colframe=vblue!65,fonttitle=\bfseries,title={#1},sharp corners,boxrule=0.6pt,left=7pt,right=7pt,top=6pt,bottom=6pt,fontupper=\small}
\title{Second-Order Problem Solving for Recursive Self-Improvement in Formal Verification}
\author{
Yuxuan Jiang$^{1,2}$\thanks{Work done during an internship at Emergence AI.}\quad
Aditya Vempaty$^{2}$\quad
Ashish Jagmohan$^{2}$\\
$^{1}$University of Maryland, Baltimore County\\
$^{2}$Emergence AI\\
\texttt{yuxuanj1@umbc.edu}
}
\iclrfinalcopy
\begin{document}
\maketitle
\fancyhead[L]{Preprint} 
\suppressfloats[t] 
\begin{abstract}

Recursive self-improvement (RSI) enables agents to iteratively optimize their workflows via execution feedback. However, standard RSI typically operates as a \emph{first-order optimizer}: it repeatedly patches surface-level parameters in response to immediate failure symptoms, often leading to trial-and-error thrashing without resolving underlying mechanisms. To address this limitation, we introduce \method{}, a framework that elevates workflow optimization to a \emph{second-order diagnostic inquiry}---investigating \emph{why} failures occur before committing to structural interventions. \method{} passively monitors execution traces for three structural anomalies (recurrence, opposing edits, and expectation mismatch) to trigger targeted mechanism investigations. By executing lightweight diagnostic probes and maintaining persistent inquiry memory across RSI rounds, \method{} accumulates causal evidence to guide systematic workflow edits rather than parameter patches. Across Lean 4 proof generation and Verus-based verifiable code generation, \method{} improves final held-out pass rates over Naive RSI by 21.8 and 25.8 percentage points under matched 24-hour search budgets. Behavioral analyses further confirm that \method{} substantially suppresses failure recurrence and eliminates unproductive zero-progress optimization loops.

\end{abstract}
\section{Introduction}\input{intro}
\section{Problem Formulation}\label{sec:problem}\input{problem_formulation}
\section{From Observed Outcomes to Supported Explanations}\label{sec:motivation}\input{pilot_findings}
\section{SO-RSI: Second-Order Problem Solving for RSI}\label{sec:method}\input{method}
\section{Empirical Evaluation}\label{sec:experiments}\input{experimental_setup}
\section{Related Work}\label{sec:related}\input{related_work}
\section{Discussion and Conclusion}\input{discussion}
\bibliography{references}
\bibliographystyle{iclr2027_conference}
\clearpage
\appendix
\section{Operational Classification of Problem-Solving Behavior}\label{app:classification}\input{appendix_classification}
\section{Additional Workflow Cases}\label{app:evidence}\input{appendix_evidence}

\section{Research Records and Evaluation Protocol}\label{app:protocol}\input{appendix_protocol}

\section{Implementation Specification}\label{app:implementation}\input{appendix_implementation}

\section{AI Use Statement}\label{app:ai-use}\input{appendix_ai_use}
\input{statements}
\clearpage
\input{related_context}
\end{document}

%% file: intro.tex
Recursive self-improvement (RSI) progressively improves an executable agent through repeated cycles of observing problems, exploring their causes, and attempting solutions \citep{robeyns2025sica}. While such quick fixes can efficiently restore progress, remaining at this level of \emph{first-order problem solving} can leave the underlying causes of those problems unexplored \citep{tucker2003hospitals}.

\begin{figure}[t]
\centering
\includegraphics[width=\linewidth]{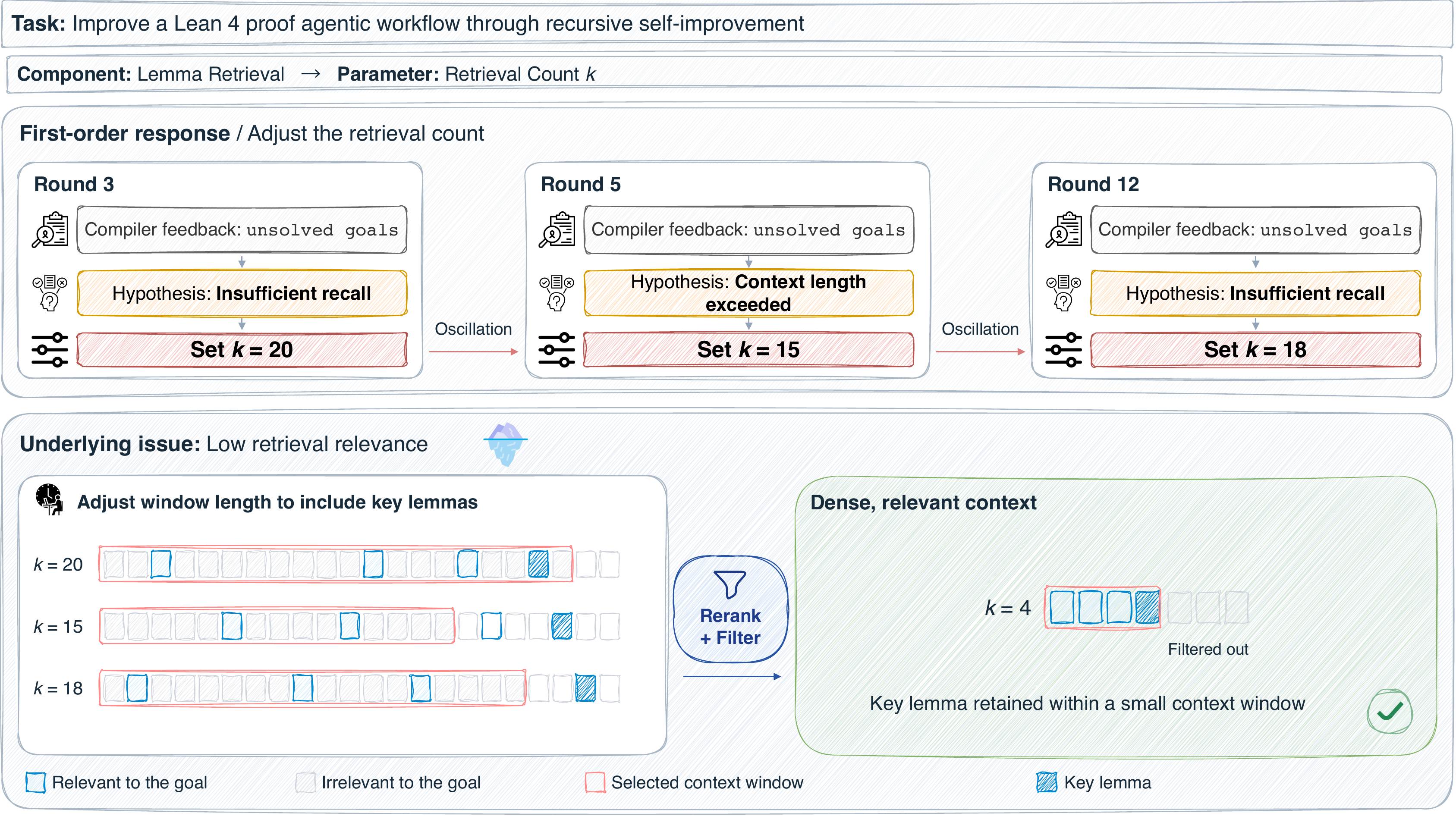}
\caption{\textbf{One symptom, opposite fixes.} Repeated \texttt{unsolved goals} feedback leads a first-order response to alternate between hypotheses of insufficient recall ($k=20$) and exceeded context limit ($k=15, 18$). \textbf{Top:} The agent oscillates through parameter patching without resolving the failure. \textbf{Bottom:} The underlying issue is poor retrieval ranking---a large window introduces excessive noise, while smaller windows exclude the low-ranked key lemma. Relevance-based reranking and filtering eliminate this trade-off, producing a dense context. Blue/dark-blue tiles denote relevant/key lemmas, and gray tiles denote noise. }
\label{fig:opening}
\end{figure}

Consider a RSI trajectory for a Lean 4 proof agentic workflow that retrieves lemmas before generating a proof (Figure~\ref{fig:opening}). When a proof attempt fails, first-order RSI relies on immediate quick fixes. Observing a failure due to missing lemmas, RSI directly increases the retrieval count to $k=20$—a quick fix that temporarily restores progress. However, a subsequent attempt fails as the excessive lemma text exceeds the context limit, prompting RSI to apply the opposite quick fix by decreasing the count to $k=15$. In a later round, as the reduced window omits critical lemmas again, RSI fine-tunes the parameter back up to $k=18$, only to encounter the original context-overflow error in the next iteration, forcing yet another reduction. This oscillation reveals a fundamental blind spot of first-order problem solving: misinterpreting a structural failure—low retrieval relevance—as a mere hyperparameter tuning task of search-testing for the "optimal lemma count," thereby failing to investigate the core mechanism of re-ranking and filtering.

This observation reveals a central risk for RSI. First-order problem solving is necessary to overcome immediate obstacles, but stopping at this level leaves the causes of recurring problems unresolved. RSI may then repeatedly confront similar failures and face conflicting prescriptions, without a clear basis for deciding which to follow. What restores progress in one attempt can therefore leave the agent caught between opposing changes in the next. Appendix~\ref{app:reversals} provides further examples involving failure history and tactic fallbacks.

Recent work on RSI and agent improvement already supports reflection, failure diagnosis, and workflow modification \citep{robeyns2025sica,chen2026failed}. The remaining question is how to direct this research when immediate fixes leave conflicting outcomes unexplained. Second-order problem solving provides a concrete target: investigate how the workflow produces the problem, obtain evidence that distinguishes explanations, and use it to guide a change to that process \citep{tucker2003hospitals}. The depth of inquiry is determined by this target, rather than by the length of reflection or the number of follow-up questions.

Our work addresses three fundamental questions. First, \textbf{first-order RSI is limited by an explanation gap}: LLMs easily propose plausible failure explanations, but without explicit verification, they resort to blind parameter patching that fails to resolve structural flaws. Second, \textbf{elevating to second-order inquiry is necessary} because stopping recurring failures requires diagnosing the underlying mechanism rather than patching surface symptoms, replacing conflicting edits with durable workflow improvements. Third, \textbf{we operationalize second-order inquiry} through a persistent diagnostic architecture that detects anomalies, formulates testable hypotheses, validates them via executable probes, and executes evidence-guided workflow edits.

We introduce \method{}, which monitors three signals: recurrence, opposing edits, and expectation mismatch. Triggered observations initiate mechanism investigation (I): gathering evidence, forming explanations, and testing predictions. Evidence accumulates across RSI rounds and guides workflow improvements (G), which return to ordinary RSI for evaluation and further optimization. Across Lean proving and Verus-based verifiable code generation (VCG), \method{} improves final pass rate over Naive RSI by 21.8 and 25.8 percentage points under the same 24-hour budget.

Our contributions are threefold:
\begin{itemize}
\item \textbf{Diagnostic analysis:} We document reversals in recorded RSI decisions and audit the evidence supporting their causal explanations.
\item \textbf{Empirical motivation:} We quantify investigation depth, intervention gains, and repeated unproductive editing across 134 attempts.
\item \textbf{The \method{} framework:} We use second-order problem solving to specify the target and evidence requirements of inquiry, integrating selective detection and persistent investigation into continuing RSI, with ablations supporting the value of probes and memory.
\end{itemize}

%% file: problem_formulation.tex
\paragraph{Two levels of iteration.}
Consider an agent attempting a formal proof or program verification task $x$. The \emph{inner loop} repairs a specific candidate artifact (e.g., a Lean 4 proof script or code annotation) given compiler feedback, keeping the governing agent procedure fixed. The \emph{outer RSI loop}---the primary focus of this work---improves the executable procedure itself, denoted as $S_t$ at round $t$. This workflow $S_t$ coordinates generation, lemma retrieval, verification, and retry policies; a \emph{workflow-level improvement} updates these reusable control components rather than the underlying language model weights. Executing $S_t$ on $x$ produces a trace $\tau_x$ of model interactions, tool outputs, and compiler feedback. An improvement agent analyzes these traces to propose a workflow edit, which is adopted into $S_{t+1}$ if validated on dev benchmarks.

\paragraph{Symptoms and mechanisms.}
A \emph{symptom} is a failure visible in $\tau_x$, such as an \texttt{unknown identifier} error. A \emph{mechanism} explains how the workflow $S_t$ produces that failure. For example, the model may have generated the required local definition, but the code extractor discarded it before compilation. The error suggests adding a definition; tracing the mechanism instead suggests repairing the extractor. This explanation is testable: compare the raw response with the submitted proof, restore the omitted definition, and recompile. We use \emph{mechanism hypothesis} for such an explanation and \emph{diagnostic probe} for a check of its predicted behavior.

\paragraph{Objective.}
Our goal is to improve final task performance within a research budget:
\begin{equation}
\max J_{\mathrm{test}}(S_T)
\quad\text{subject to}\quad
\sum_{t=0}^{T-1} C_t \leq B,
\label{eq:objective}
\end{equation}
where $J_{\mathrm{test}}$ is held-out task success and $C_t$ includes investigation, editing, and evaluation costs at round $t$. Only development tasks guide improvement. We report token use, verifier calls, and wall time, and track failure recurrence to assess whether an intervention yields durable progress.

%% file: pilot_findings.tex
RSI can often make progress through first-order problem solving: observe a failure, apply a plausible fix, and continue. But when is such a quick fix enough? A local intervention may resolve the current failure while leaving unexplained why it occurred, allowing the same underlying problem to reappear or invite a later, opposing fix. This motivates a second-order question: does the agent investigate the workflow mechanism that generates the failure before continuing to optimize it?

Inspired by the distinction between immediate workarounds and efforts to address underlying causes in \citet{tucker2003hospitals}, we operationalize this transition for RSI as a progression in observable inquiry. F (direct repair) applies an immediate fix without a workflow-mechanism explanation. H (mechanism hypothesis) proposes such an explanation but does not test it. I (mechanism investigation) performs a diagnostic check of the proposed mechanism. G (evidence-guided intervention) uses the resulting evidence to determine the intervention. U (insufficient information) marks attempts whose inquiry process cannot be determined from the available record. For a conservative binary summary, only F is counted as first-order; H, I, and G represent progressively stronger observable second-order inquiry, while U is grouped on the second-order side only to avoid inflating the first-order count. Appendix~\ref{app:classification} provides the full decision rules and worked examples.

To examine the gap between reactive patching and structural diagnosis, we audit five independent Lean 4 proof RSI trajectories comprising 134 optimization attempts. We ask two research questions:
\begin{itemize}[leftmargin=*,noitemsep,topsep=2pt]
    \item \textbf{RQ1:} How often does RSI move beyond direct patching?
    \item \textbf{RQ2:} What gains does RSI achieve, and when does further patching become unproductive?
\end{itemize}

\noindent\textbf{RQ1: How often does RSI move beyond direct patching?}

Our analysis covers all 134 optimization attempts across the five trajectories and uses each complete attempt as its unit, including the observation, diagnosis, checks, and intervention. A change to a shared parameter is classified by the inquiry behind it, not by the scope of the edit or its observed gain.

Table~\ref{tab:inquiry} nests the five inquiry labels under the binary grouping. Direct repair accounts for 86 of 134 attempts (64.2\%). Although 48 attempts (35.8\%) fall on the permissive second-order side, 29 contain only an untested mechanism hypothesis and four have insufficient records. Only 15 attempts (11.2\%) include an observable diagnostic check: eight investigations (I) and seven evidence-guided interventions (G). Among the 44 attempts that articulate a mechanism (H/I/G), only 15 (34.1\%) test it. The main gap is therefore not an absence of explanations, but a failure to turn explanations into diagnostic checks. All percentages use the 134 outer optimization attempts as their denominator; inner proof-repair calls are not separate units.

\begin{table}[t]
\caption{\textbf{Problem-solving behavior in outer RSI (RQ1).} Under our permissive convention, only direct repairs (\textbf{F}) count as first-order. The second-order group contains hypotheses (\textbf{H}), diagnostic investigations (\textbf{I}), evidence-guided interventions (\textbf{G}), and insufficient records (\textbf{U}). I+G denotes an observable diagnostic check, not necessarily a confirmed mechanism or successful intervention. Shares use all 134 attempts across five trajectories.}
\label{tab:inquiry}
\centering\small
\begin{tabularx}{\linewidth}{@{}Yrr@{}}
\toprule
\textbf{Inquiry category} & \textbf{Attempts} & \textbf{Share} \\
\midrule
\textbf{First-Order: Direct repair (F)} & \textbf{86} & \textbf{64.2\%} \\
\midrule
\textbf{Second-Order Behavior (Permissive Grouping)} & \textbf{48} & \textbf{35.8\%} \\
\quad Mechanism hypothesis, untested (\textbf{H}) & 29 & 21.6\% \\
\quad Mechanism investigation (\textbf{I}) & 8 & 6.0\% \\
\quad Evidence-guided intervention (\textbf{G}) & 7 & 5.2\% \\
\quad Insufficient information (\textbf{U}) & 4 & 3.0\% \\
\midrule
\textbf{Total attempts} & \textbf{134} & \textbf{100.0\%} \\
\bottomrule
\end{tabularx}
\end{table}

\noindent\textbf{RQ2: What gains does RSI achieve, and when does further patching become unproductive?}

RSI improves proof success across all five trajectories. Table~\ref{tab:rsi-gains-thrashing} shows that direct repair delivers useful gains, while diagnostic inquiry (I/G) achieves the largest average improvement per evaluated edit. Regressions occur in every group, highlighting the need to distinguish productive repairs from edits that warrant further investigation.

Repeated work is concentrated in direct repair: eight of the nine zero-progress loops assigned to these groups contain only F edits, repeatedly changing the workflow without retaining a performance or cost benefit. Together with the scarcity of diagnostic checks in RQ1, this finding motivates \emph{selective escalation}: retain efficient direct repairs, but investigate the generating mechanism when repeated patches fail to produce durable progress. Appendix~\ref{app:rq2-details} provides the detailed results and analysis protocol.

\begin{table}[t]
\centering\small
\caption{\textbf{Gains and regressions of evaluated RSI edits (RQ2).} Gain is the pass-rate change relative to the edit's actual parent, in percentage points (pp); regressions are edits with a negative change. Results cover 79 evaluated edits from 130 F/H/I/G attempts; arrows indicate the preferred direction.}
\label{tab:rsi-gains-thrashing}
\begin{tabularx}{\linewidth}{@{}Yrrr@{}}
\toprule
\textbf{Inquiry group} & \shortstack{\textbf{Evaluated}\\\textbf{edits}} & \shortstack{\textbf{Mean gain} \textcolor{vgreen}{$\uparrow$}\\\textbf{(pp)}} & \shortstack{\textbf{Performance}\\\textbf{regressions} \textcolor{vblue}{$\downarrow$}} \\
\midrule
Direct repair (F) & 51 & +8.7 & 20/51 (39.2\%) \\
Untested hypothesis (H) & 17 & +11.2 & 4/17 (\textbf{23.5\%}) \\
Diagnostic inquiry (I/G) & 11 & \textbf{+14.3} & 3/11 (27.3\%) \\
\bottomrule
\end{tabularx}
\end{table}

%% file: method.tex
\begin{figure}[t]
\centering\includegraphics[width=\linewidth]{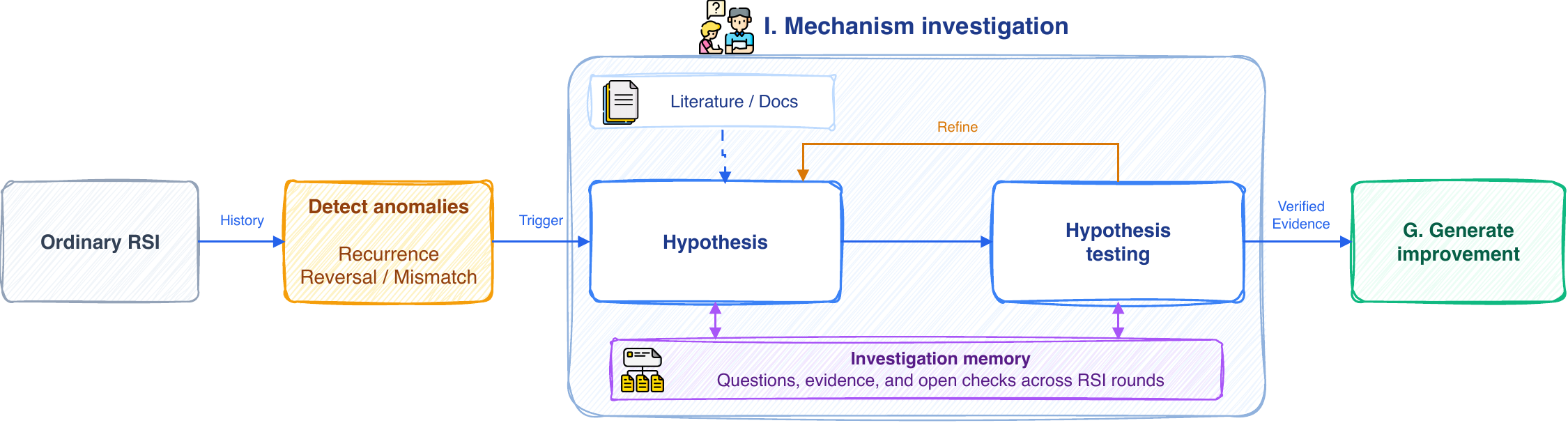}
\caption{\textbf{From first-order thrashing to second-order improvement.} Anomalies in the RSI history trigger mechanism investigation (I); diagnostic evidence guides a workflow change (G), evaluated by ordinary RSI. Within I, the agent forms and tests hypotheses, consults external knowledge when needed, and retains findings across RSI rounds.}
\label{fig:method}
\end{figure}

\method{} detects signs of first-order thrashing and redirects RSI toward investigating the mechanisms behind them (Figure~\ref{fig:method}). A \textbf{lightweight detector} (\S\ref{sec:questions}) identifies improvement anomalies from execution logs. A \textbf{mechanism questioner (I)} (\S\ref{sec:investigation}) then gathers evidence, formulates hypotheses, and executes diagnostic probes to test them. The resulting diagnostic evidence guides an \textbf{evidence-guided intervention (G)} (\S\ref{sec:construction}), which translates the investigated causes into systematic workflow changes. This makes the improvement second-order: it addresses the process producing the observed failures, rather than repeatedly patching their symptoms. The revised workflow then returns to ordinary RSI for evaluation.

 Appendix~\ref{app:implementation} describes the controller and investigation interfaces.

\subsection{Lightweight Detection of Improvement Anomalies}\label{sec:questions}
After each RSI optimization attempt, the detector logs a structured record $X_t=\{v_t,d_t,r_t\}$, where $v_t$ is the workflow version, $d_t$ specifies the modified component and textual edit, and $r_t$ captures the intended rationale. Rather than running additional benchmark evaluations, the detector filters prior records by component or failure type, then uses a short LLM call to compare $X_t$ with the selected traces $\tau_x$ for three structural \emph{improvement anomalies}:
\textbf{recurrence}, where targeted failure signatures persist or re-emerge across consecutive rounds despite attempted edits; \textbf{reversal}, where edges or parameters are directly toggled or opposingly altered for similar symptoms; and \textbf{expectation mismatch}, where observed outcomes directly contradict the stated rationale $r_t$ or predicted behavior. When any anomaly is detected, $X_t$ and its context are routed to initiate or update a mechanism investigation; absent any signal, ordinary RSI continues without investigation. Detection calls count toward the research budget.

\subsection{Mechanism Investigation (I)}\label{sec:investigation}
Given a flagged record $X_t$, the system investigates the root cause before committing to permanent workflow edits. Investigation operates via a two-stage diagnostic loop: \emph{hypothesis formation} and \emph{hypothesis testing}.

\paragraph{Hypothesis formation.}
The questioner first retrieves relevant historical attempts to contrast trajectories. For instance, observing a successful attempt with fewer lemmas directly challenges the naive assumption that retrieval count is the primary bottleneck. The questioner then analyzes workflow interactions to propose non-mutually-exclusive causal hypotheses (e.g., failure to retrieve vs. suboptimal ranking). Each hypothesis specifies the implicated workflow operation, supporting or conflicting observations, a testable prediction, and a proposed check. When domain mechanics or diagnostic methods are uncertain, it consults external documentation or code specs on demand to ground its hypotheses.

\paragraph{Hypothesis testing (Diagnostic Probing).}
The questioner generates a targeted diagnostic script or tool call, which the controller executes in an isolated workspace. It first inspects existing artifacts before requesting new generations. For example, saved retrieval lists distinguish missing lemmas from low-ranked ones; comparing original and reranked contexts under fixed context budgets and generator settings then tests the effect of ranking. The agent compares actual outputs with the recorded prediction. A completed check supporting an actionable workflow change permits handoff to G; otherwise the hypothesis is revised or retained for later investigation.

\paragraph{Persistent investigation memory.}
Inspired by experience maintenance in AutoManual~\citep{chen2024automanual}, we maintain an investigation record containing linked attempts, hypotheses, executed checks, their results, and the next open question. Related anomalies update the same investigation, allowing later RSI rounds to resume from the evidence already collected.

\subsection{Evidence-Guided Improvement (G)}\label{sec:construction}
G reuses the ordinary RSI editing agent, supplying the supported explanation, probe results, target code, and expected behavioral change. Its output is a workflow patch, which may adjust an existing component or introduce a new one. For the retrieval example, evidence of ranking failure motivates reranking and filtering before context construction.

The candidate revision returns to the outer RSI loop for validation. Only accepted edits update $S_{t+1}$; all evaluation results feed back into investigation memory. Subsequent RSI can further optimize accepted changes under the shared budget in Equation~\ref{eq:objective}.

%% file: experimental_setup.tex
\subsection{Experimental Setup}
\paragraph{Tasks and data.}
For Lean 4 proof generation, we adopt the initial workflow and optimization setting from RSI-Exam~\citep{rsiexam2026tasks}, retaining its original problems and supplementing them with problems from VeriSoftBench~\citep{xin2026verisoftbench} to obtain 92 training and 48 test problems. Each outer RSI run has a 24-hour budget. For Verus-based verifiable code generation (VCG), we construct an analogous agentic workflow using 113 training and 46 test problems labeled \texttt{hard} in VeriContest~\citep{xie2026vericontest}, with the same 24-hour budget. GPT-5.6-sol performs outer RSI optimization; the inner task-solving model and per-task budgets follow RSI-Exam. We report means and standard deviations over three independent runs. Task details are provided in Appendix~\ref{app:task-settings}.

\paragraph{Initial workflow and compared methods.}
The seed workflow $S_0$ is the shared starting point for RSI: it generates a candidate, checks it with the verifier, and, upon failure, performs one feedback-guided repair followed by another check. For Lean, candidates are proof bodies; for VCG, they are Verus program artifacts. \textbf{Naive RSI} improves this workflow through the standard feedback-and-edit loop. \textbf{\method{}} augments the same loop with anomaly detection, persistent investigation, and evidence-guided intervention.

\subsection{Overall Performance and Improvement Trajectories}
\begin{figure}[t]
\centering
\includegraphics[width=\linewidth]{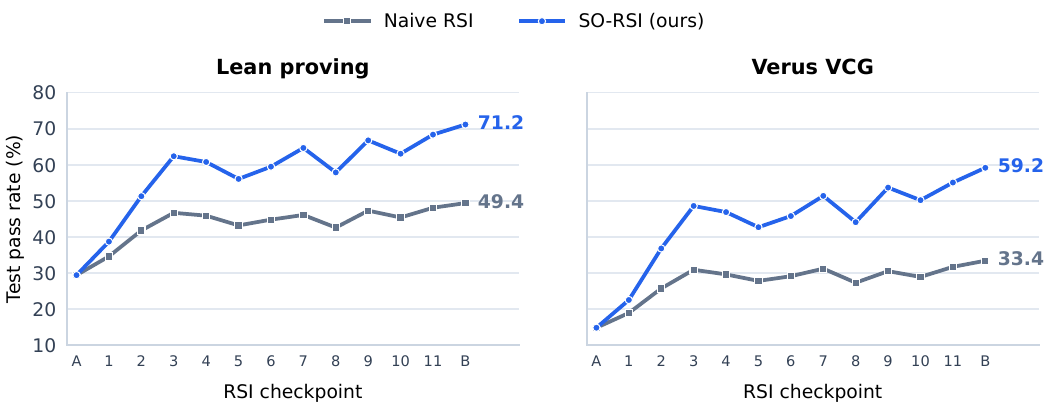}
\caption{\textbf{Verification performance across RSI checkpoints.} Naive RSI and \method{} on Lean proving and Verus VCG. A denotes the initial workflow and B the final result; 1--11 are intermediate checkpoints shown in sequence, not elapsed-time intervals. Curves show the reported pass rates without checkpoint-level uncertainty estimates.}
\label{fig:budget-performance}
\end{figure}

The primary metric is held-out task pass rate. We also report its change from the shared seed in percentage points (pp). Table~\ref{tab:main-results} reports each domain separately, reporting means and standard deviations across three independent outer runs. Figure~\ref{fig:budget-performance} compares pass rate across the reported RSI checkpoints from the initial workflow to the final result. Checkpoint workflows are frozen during RSI and tested afterward; test outcomes never guide optimization.

\paragraph{Higher final performance under the same time budget.}
\method{} reaches 71.2\% on Lean and 59.2\% on VCG, exceeding Naive RSI by 21.8 and 25.8 percentage points, respectively (Table~\ref{tab:main-results}). Naive RSI itself improves substantially over the seed, so the advantage is additional to the gains from ordinary workflow optimization. Figure~\ref{fig:budget-performance} shows that \method{} remains ahead at every reported checkpoint after initialization, although both methods exhibit temporary regressions. Second-order investigation improves the final outcome without eliminating all unsuccessful intermediate changes.

\begin{table}[t]
\centering\small
\setlength{\tabcolsep}{5pt}
\caption{\textbf{Held-out verification performance.} RSI methods share a 24-hour outer budget and identical per-task limits. Pass is task pass rate (\%); $\Delta$ is the change from the corresponding seed (pp). Pass reports mean $\pm$ standard deviation over three independent runs; $\Delta$ is the mean gain over the fixed starting-point reference $S_0$.}
\label{tab:main-results}
\begin{tabularx}{\linewidth}{@{}Yrrrr@{}}
\toprule
& \multicolumn{2}{c}{\textbf{Lean proving}} & \multicolumn{2}{c}{\textbf{Verus VCG}} \\
\cmidrule(lr){2-3}\cmidrule(l){4-5}
\textbf{Method} & \textbf{Pass $\uparrow$} & \textbf{$\Delta\uparrow$} & \textbf{Pass $\uparrow$} & \textbf{$\Delta\uparrow$} \\
\midrule
Initial workflow $S_0$ & 29.4 & 0.0 & 14.8 & 0.0 \\
Naive RSI & $49.4\pm2.8$ & +20.0 & $33.4\pm5.2$ & +18.6 \\
\method{} & $71.2\pm1.9$ & +41.8 & $59.2\pm3.7$ & +44.4 \\
\bottomrule
\end{tabularx}
\end{table}

\section{Ablation Study}
Two ablations test the central design choices. \textbf{Without executable probes} retains hypothesis formation and textual assessment but removes diagnostic execution, testing the value of checking explanations against actual behavior. \textbf{Without persistent investigation memory} clears investigation records between outer rounds while preserving accepted workflow edits and ordinary RSI logs, testing the value of accumulating inquiry across attempts. Each uses the same overall research budget as the full method. Table~\ref{tab:method-ablations} reports final held-out pass rate for both domains.

\begin{table}[t]
\centering\small
\caption{\textbf{Contribution of executable probes and persistent investigation memory.} Held-out pass rates (\%) are reported as mean $\pm$ standard deviation over three runs. All variants share the initial workflow and 24-hour outer budget.}
\label{tab:method-ablations}
\begin{tabularx}{\linewidth}{@{}Yrr@{}}
\toprule
\textbf{Variant} & \textbf{Lean pass $\uparrow$} & \textbf{VCG pass $\uparrow$} \\
\midrule
\method{} & $71.2\pm1.9$ & $59.2\pm3.7$ \\
\quad w/o executable probes & $63.4\pm5.6$ & $33.0\pm3.8$ \\
\quad w/o persistent investigation memory & $60.9\pm4.2$ & $36.7\pm6.2$ \\
\bottomrule
\end{tabularx}
\end{table}

\paragraph{Both diagnostic execution and memory contribute.}
Removing executable probes lowers final pass rate by 7.8 points on Lean and 26.2 points on VCG; removing persistent investigation memory lowers it by 10.3 and 22.5 points (Table~\ref{tab:method-ablations}). Textual hypothesis formation alone does not recover the full benefit, and restarting investigations each round also weakens improvement. Both ablations still outperform Naive RSI on Lean, whereas the VCG probe ablation falls close to the baseline. These results support combining executable checks with evidence retained across rounds.

\subsection{Analysis of Improvement Behavior}
We examine whether investigation changes how RSI improves its workflow, comparing Naive RSI, \method{}, and both ablations. We apply the behavioral definitions from Section~\ref{sec:motivation} to the actual records: invoking an I/G module does not itself establish second-order behavior. The same criteria capture spontaneous investigation in Naive RSI.

\paragraph{Investigation changes improvement decisions.}
Table~\ref{tab:second-order-behavior} shows that \method{} tests mechanism explanations in 46.0\% of Lean investigations and 41.6\% of VCG investigations, compared with 14.8\% and 8.9\% for Naive RSI. Evidence-driven decision revisions also become more frequent, reaching 22.6\% and 20.6\%. These revisions can redirect an intervention or stop an unsupported proposal; confirming an initial explanation remains useful as well. The accompanying G edits have higher acceptance rates and larger mean development gains: 9.8 versus 4.4 points on Lean, and 12.0 versus 5.2 points on VCG.

\paragraph{Improvements reduce recurrence and repeated editing.}
On the audited cases, \method{} reduces mechanism recurrence by 31.4 points on Lean and 35.7 points on VCG, compared with 7.3 and 10.9 points for Naive RSI. Zero-progress loops fall from 18.7 to 7.2 and from 21.4 to 8.6 per 100 evaluated edits, respectively. Removing either probes or memory weakens both outcomes. In particular, the memory ablation still executes many diagnostic checks but revises fewer decisions and achieves smaller recurrence reductions, suggesting that retaining and connecting evidence matters alongside executing checks. Recurrence refers to the diagnosed mechanism, rather than repeated compiler wording; Appendix~\ref{app:second-order-analysis} specifies the audit and aggregation rules.

\begin{table}[t]
\centering\scriptsize
\setlength{\tabcolsep}{3pt}
\caption{\textbf{How investigation translates into workflow improvement.} The first three columns are percentages calculated from their respective reported numerator/denominator counts (Appendix~\ref{app:behavior-counts}); Tested and Revised share the denominator of all investigation episodes; Revised counts changes driven by executed diagnostic evidence. G acceptance is a percentage of evaluated G edits. $\Delta$Pass is the mean parent-to-edit development gain (pp), including rejected edits. $\Delta$Rec. is post-edit minus pre-edit mechanism recurrence (pp; negative indicates reduction). Loops are zero-progress loops per 100 evaluated edits. Dashes in the w/o probes row mark probe-dependent metrics as not applicable. The table reports point estimates; run-level standard deviations are not available for these behavioral metrics.}
\label{tab:second-order-behavior}
\begin{tabularx}{\linewidth}{@{}Yrrrrrr@{}}
\toprule
& \multicolumn{2}{c}{\textbf{Investigation}} & \multicolumn{2}{c}{\textbf{G edits}} & \multicolumn{2}{c}{\textbf{Persistence}} \\
\cmidrule(lr){2-3}\cmidrule(lr){4-5}\cmidrule(l){6-7}
\textbf{Method} & \textbf{Tested (\%)} & \textbf{Revised (\%)} & \textbf{Accept (\%)} & \textbf{$\Delta$Pass $\uparrow$} & \textbf{$\Delta$Rec. $\downarrow$} & \textbf{Loops $\downarrow$} \\
\midrule
\multicolumn{7}{l}{\textit{Lean proving}} \\
Naive RSI & 14.8 & 0.9 & 54.8 & 4.4 & -7.3 & 18.7 \\
\method{} & 46.0 & 22.6 & 67.1 & 9.8 & -31.4 & 7.2 \\
\quad w/o probes & --- & --- & 56.5 & 6.1 & -12.8 & 13.9 \\
\quad w/o memory & 41.6 & 15.3 & 59.7 & 7.0 & -19.6 & 11.8 \\
\midrule
\multicolumn{7}{l}{\textit{Verus VCG}} \\
Naive RSI & 8.9 & 3.4 & 53.4 & 5.2 & -10.9 & 21.4 \\
\method{} & 41.6 & 20.6 & 68.1 & 12.0 & -35.7 & 8.6 \\
\quad w/o probes & --- & --- & 56.5 & 7.4 & -14.5 & 16.2 \\
\quad w/o memory & 37.7 & 14.9 & 60.0 & 8.3 & -23.1 & 13.7 \\
\bottomrule
\end{tabularx}
\end{table}

\paragraph{What explains the gains?}
The results connect more frequent mechanism testing with better intervention outcomes and less repeated work. Naive RSI already performs some diagnostic inquiry and reduces recurrence, but \method{} makes this behavior more systematic. Together with the ablations, the evidence supports selective second-order investigation as a complement to direct repair: it helps RSI revise its explanation when needed and carry that evidence into reusable workflow changes.

%% file: related_work.tex
\subsection{Recursive Self-Improvement and Workflow Diagnosis}
SICA improves a coding agent by letting it edit its own implementation and evaluate the resulting changes~\citep{robeyns2025sica}. HarnessFix more directly connects failed trajectories to harness mechanisms, consolidates diagnoses, and validates targeted repairs~\citep{chen2026failed}. These works establish self-modification and mechanism diagnosis as existing capabilities. SO-RSI studies when repeated improvement attempts call for further investigation: recurring failures, opposing edits, and unmet expectations trigger an inquiry that persists across RSI rounds and tests explanations before guiding another workflow edit.

\subsection{Second-Order Problem Solving}
\citet{tucker2003hospitals} distinguish immediate workarounds from efforts to address the processes that produce recurring failures. We adapt this distinction to outer RSI, using recorded hypotheses, diagnostic checks, and evidence-guided interventions to characterize inquiry. The distinction concerns what an agent investigates and tests, rather than the size of its edit. SO-RSI retains ordinary repairs while selectively investigating the workflow mechanisms behind unresolved patterns. A related distinction separates comprehension failures from faulty reasoning, motivating changes to how a model reads a question~\citep{han2026read}; our investigation concerns the reusable workflow that produces failures across attempts.

\subsection{Agentic Formal Verification}
LeanSearch v2 improves Lean proving through global premise retrieval, while RAG-Verus retrieves repository context for program verification~\citep{gao2026leansearch,zhong2025rag}. ExVerus generates and validates counterexamples to guide Verus proof repair~\citep{yang2026exverus}. These methods provide concrete ways to improve the information available to a solver. In code security, behavior-aware extraction and demonstration selection likewise organize the context supplied to an LLM detector~\citep{han2025can}. SO-RSI addresses the outer process that chooses such changes: it uses diagnostic evidence to determine which workflow mechanism to revise. Retrieval and counterexample reasoning are possible interventions, rather than components prescribed in advance by our method.

\paragraph{Context-sensitive diagnosis and guidance.}
\citet{xu2026chainawareencodingmicroservicetrace} condition microservice anomaly detection on invocation chains, exposing path failures that appear normal when events are considered separately.
SO-RSI similarly considers the context of an observed failure, but investigates patterns across workflow edits rather than predicting anomalous service events.
\citet{li2026gracegroundedreasoningadapter} compose lightweight adapters using educational state and calibrate answer options against multimodal evidence.
Their task-specific adaptation is distinct from our outer-loop investigation, which determines which reusable workflow mechanism needs to change.
\citet{li2026complementary} find that more elaborate AI-generated theories do not necessarily predict empirical results better; SO-RSI consequently tests mechanism explanations rather than rewarding their complexity.
Broader comparisons with reasoning supervision and task-grounded evaluation appear in Appendix~\ref{app:related-context}.

%% file: discussion.tex
We studied second-order problem solving for outer RSI in Lean proof generation and Verus-based code verification. \method{} detects improvement anomalies, maintains investigations across rounds, and uses executable diagnostic evidence to guide workflow interventions. Under the same 24-hour outer budget, it improves final held-out pass rate over Naive RSI by 21.8 percentage points on Lean and 25.8 points on VCG. Ablations show that both executable probes and persistent investigation memory contribute to these gains.

The behavioral analysis links these gains to more frequent mechanism testing, more evidence-driven revisions, and stronger intervention outcomes. Diagnosed failures recur less often, and zero-progress loops decline in both domains. The aim is selective escalation: direct repairs remain useful, while recurring or conflicting outcomes warrant investigation of the workflow mechanism that produces them.

\paragraph{Limitations.}
Our evaluation covers two formal-verification domains and one outer improvement model. Behavioral findings concern the recorded trajectories and audited mechanisms; generalization to other models and workflows remains an open question.

%% file: appendix_classification.tex
We operationalize the distinction inspired by \citet{tucker2003hospitals} for RSI; it is not a verbatim reproduction of their organizational-behavior coding scheme. The corpus contains five independent Lean 4 proof RSI trajectories and 134 optimization attempts. The annotation unit is a complete optimization attempt, including its observation, diagnosis, checks, and intervention. Neither changing a shared parameter nor adding a component determines the label. Each label must be linked to recorded evidence; unrecorded reasoning is not reconstructed from the eventual outcome.

\paragraph{Diagnostic criteria.}
Table~\ref{tab:order-criteria} separates testing an intervention's performance from testing an explanation of failure. A diagnostic check may inspect code or traces, replay an artifact, or execute a controlled probe. It need not be a new model call or succeed in confirming the hypothesis.

\begin{table}[htbp]
\centering\small
\caption{Observable criteria for mechanism-directed investigation. These criteria describe a progression, not an additive score.}
\label{tab:order-criteria}
\begin{tabularx}{\linewidth}{@{}p{0.18\linewidth}YY@{}}
\toprule
\textbf{Criterion} & \textbf{Evidence required} & \textbf{Insufficient evidence} \\
\midrule
M: Mechanism explanation & Identifies how a workflow process produces the failure, beyond restating the symptom. & ``Too few lemmas; increase $k$'' merely redescribes a shortage. \\
\addlinespace
D: Diagnostic check & Actually checks whether the proposed mechanism occurs or distinguishes competing causes; the target is not only aggregate task performance. & Comparing solve rates at $k=10$ and $k=20$ alone tests settings, not the cause of failure. \\
\addlinespace
L: Evidence-guided intervention & The intervention is explicitly justified by diagnostic findings; contradictory or inconclusive evidence can change or halt the proposal. & Increasing $k$ regardless of the diagnostic result. \\
\bottomrule
\end{tabularx}
\end{table}

\paragraph{Fine-grained labels and binary aggregation.}
We use a lenient criterion for crediting second-order behavior, thereby conservatively limiting the first-order count: only F counts as first-order; H, I, G, and U all count on the second-order side. U retains its meaning of insufficient observability. Its inclusion is an explicit conservative accounting convention, not evidence of second-order reasoning. We retain all five labels and their counts so that the binary aggregate does not obscure this distinction. In particular, the second-order share under this convention must not be described as the share of validated mechanism investigations.

\begin{table}[htbp]
\centering\small
\caption{Fine-grained labels and the permissive binary convention. Lack of diagnostic evidence yields H when a mechanism hypothesis is recorded; inadequate records to judge the attempt yield U.}
\label{tab:order-labels}
\begin{tabularx}{\linewidth}{@{}p{0.21\linewidth}Yp{0.17\linewidth}@{}}
\toprule
\textbf{Label} & \textbf{Decision rule} & \textbf{Binary group} \\
\midrule
F: Direct repair & Observable behavior is limited to immediate repair or settings search, without an upstream mechanism explanation. & First-order \\
\addlinespace
H: Mechanism hypothesis & M is present, but no diagnostic check D is carried out. & Second-order \\
\addlinespace
I: Mechanism investigation & M is present and D is carried out; findings may be inconclusive or refute the hypothesis. & Second-order \\
\addlinespace
G: Evidence-guided intervention & M and D are present, and the resulting intervention satisfies L. Task-level success is not required for this label. & Second-order \\
\addlinespace
U: Insufficient information & The available record does not support a judgment of the attempt's inquiry process. & Second-order, by convention \\
\bottomrule
\end{tabularx}
\end{table}

\paragraph{Worked examples.}
Table~\ref{tab:order-examples} uses the illustrative lemma-retrieval scenario. A gain from changing $k$ does not by itself establish that missing evidence caused the failure. Conversely, a diagnostic investigation may justify changing only $k$; intervention complexity is not a classification criterion.

\begin{table}[htbp]
\centering\small
\caption{Illustrative annotations for lemma retrieval. These are coding examples, not measured experimental trajectories.}
\label{tab:order-examples}
\begin{tabularx}{\linewidth}{@{}YcY@{}}
\toprule
\textbf{Observed optimization behavior} & \textbf{Label} & \textbf{Rationale} \\
\midrule
Claims that more lemmas are needed; raises $k$ from 10 to 20 and observes a higher solve rate. & F & Establishes a setting's performance, not why the failure occurred. \\
\addlinespace
Later claims that there are too many lemmas; lowers $k$ from 20 to 15 and observes a higher solve rate. & F & Opposite settings search; evaluation across instances does not itself establish mechanism inquiry. \\
\addlinespace
Hypothesizes that lexical ranking pushes a key lemma below top-$k$, then replaces the retriever without checking that explanation. & H & Provides a process-level hypothesis but no diagnostic check. \\
\addlinespace
Checks the needed lemma's rank in failed cases and whether it actually entered the generation prompt. & I & Tests whether key evidence was omitted by the retrieval process. \\
\addlinespace
Holds the count fixed while replacing an irrelevant lemma with a needed one; or preserves needed lemmas while removing irrelevant ones. & I & Probes evidence quality or distraction rather than only searching over counts. \\
\addlinespace
Uses diagnostic findings to select ranking, filtering, or count changes, then evaluates the intervention. & G & The intervention follows from mechanism evidence rather than a predetermined solution. \\
\bottomrule
\end{tabularx}
\end{table}

\paragraph{Annotation record and reporting.}
For each attempt, record the problem, mechanism hypothesis, diagnostic action and result, intervention rationale, label, and exact trace references. Multiple functional measures within an attempt may receive subsidiary annotations but should not be treated as independent attempts. Reviewers should independently code an initial sample, resolve ambiguities in the rules, and report agreement before adjudication. RQ1 and RQ2 share the five-trajectory, 134-attempt scope. Inner repair calls and outer optimization attempts have different units and must not be pooled into one proportion.

\begin{table}[htbp]
\centering\small
\caption{\textbf{Fine-grained breakdown underlying RQ1.} All shares use the same denominator of 134 outer optimization attempts across five trajectories as Table~\ref{tab:inquiry}. The subtotal I+G identifies attempts with an observable diagnostic check, including inconclusive or refuted hypotheses. U is retained explicitly despite its inclusion on the second-order side. Percentages are rounded to one decimal place.}
\label{tab:order-counts}
\begin{tabularx}{\linewidth}{@{}Yrr@{}}
\toprule
\textbf{Label} & \textbf{Attempts} & \textbf{Share} \\
\midrule
F: Direct repair & 86 & 64.2\% \\
H: Mechanism hypothesis & 29 & 21.6\% \\
I: Mechanism investigation & 8 & 6.0\% \\
G: Evidence-guided intervention & 7 & 5.2\% \\
U: Insufficient information & 4 & 3.0\% \\
\midrule
Total & 134 & 100\% \\
\addlinespace
Diagnostic investigation subtotal (I+G) & 15 & 11.2\% \\
\bottomrule
\end{tabularx}
\end{table}

\paragraph{Thrashing attribution.}
For a numeric setting, a reversal occurs when consecutive nonzero changes satisfy $(k_t-k_{t-1})(k_{t-1}-k_{t-2})<0$. The reversal rate is the number of reversals divided by eligible adjacent edit pairs, whereas a category's share is its fraction of all reversals. Counts use the five inquiry labels and retain mixed attribution separately. Reversals from different trajectories or branches are not concatenated.

A component/parameter key or a stable prompt-section key identifies each edit location. Numeric deltas are evaluated within an actual parent--child lineage, not by concatenating trials from different branches. Shared lineage prefixes are counted once. Prompt oppositions are supported by before/after instructions and trace references. A reversal is attributed to the closing edit; mixed attempts retain mixed attribution. Claims about the share of reversals attributable to F use all recorded reversals as the denominator, including unknown and mixed cases. Zero-progress loops are assessed separately using comparable task and cost evaluations; a shorter accepted proof, a lower-cost workflow, or an evidence-justified adaptation is not automatically a zero-progress loop. Overlapping returns within the same excursion are not counted as independent loops. The observation horizon and number of available edit opportunities accompany comparisons between groups.

\paragraph{Overall gains and retained progress.}
Each trajectory's overall gain compares its initial workflow with the final workflow selected by that run's acceptance rule on the same task set. The highest observed trial score is not substituted for the final accepted score; any best-trial statistic is reported separately. All 134 optimization attempts contribute to their trajectories' recorded research costs, including investigations and rejected trials. Cost coverage includes improvement-agent and task-model calls where recorded; incomplete accounting is reported explicitly rather than treated as zero. Overall gains are not obtained by summing trial deltas and are not causally partitioned among inquiry labels, because accepted descendants inherit and interact with earlier edits.

Retained progress is assessed on comparable descendant evaluations of an accepted intervention against its original parent, using both task success and model-call or token costs. Reports include the follow-up horizon and eligible opportunities; lack of subsequent exposure or incomplete follow-up yields an unknown outcome, not evidence of durable success. Reversal counts remain separate from zero-progress-loop counts. Costs assigned to overlapping loops are deduplicated by optimization-attempt identifier and reported alongside the total research cost. A failed direct patch does not by itself establish that a diagnostic investigation was necessary.

The RQ2 comparisons use three mutually exclusive groups: F, H, and I/G. U is included in trajectory-level totals but excluded from the three-group comparison. A loop is assigned to a group only when all its constituent edits belong to that group; loops spanning groups or involving U remain separately recorded and are not assigned to any displayed group. Thus, displayed loop counts need not sum to the total number of loops. Reversals use the closing edit's group; unknown or mixed attribution remains separately recorded. The underlying record retains the original fine-grained labels. Group comparisons report eligible evaluation counts, and investigations without an evaluated workflow change are not counted as ties. Event-count comparisons are descriptive and must account for the unequal group sizes and available edit opportunities. Edit-level mean pass-rate change (percentage points) and Up/Tie/Down summaries use all eligible, fully evaluated attempts from the five trajectories, each compared with its actual parent. Evaluation coverage and non-evaluable attempts are reported explicitly. Net solve-count changes are interpreted within task sets; percentage-point changes accompany pooled comparisons if task-set sizes differ.

\subsection{Detailed RQ2 Results}\label{app:rq2-details}

Table~\ref{tab:rq2-details} preserves the trajectory-level results and full edit-level statistics behind the compact main-text comparison. Comparisons are descriptive: group sizes and evaluation coverage differ, and later workflows inherit earlier edits. These statistics do not isolate the causal effect of inquiry depth. Performance regression rates are 20/51 (39.2\%) for F, 4/17 (23.5\%) for H, and 3/11 (27.3\%) for I/G. A reversal is not necessarily harmful: it may reflect a justified adaptation. Zero-progress loops additionally require the absence of retained performance or cost benefit. The eight F-only loops out of nine assigned loops describe the composition of those events, not their incidence per editing opportunity.

\begin{table}[t]
\centering\small
\setlength{\tabcolsep}{4pt}
\caption{\textbf{Detailed trajectory and edit outcomes for RQ2.} Top: pass rates (\%), final-minus-initial gain (percentage points), and total cost (million tokens). Bottom: three mutually exclusive inquiry groups; $\Delta$ is the pass-rate change in percentage points relative to an attempt's actual parent. U remains in trajectory totals but is excluded from group comparisons. Reversals use the closing edit's group; loops spanning groups are not assigned to a single group. The summary row averages pass rates and gains and sums costs.}
\label{tab:rq2-details}
\begin{tabularx}{\linewidth}{@{}Yrrrr@{}}
\toprule
\textbf{Trajectory} & \textbf{Initial} & \textbf{Final} & \textbf{$\Delta$} & \textbf{Cost} \\
\midrule
1 & 29.4 & 46.2 & +16.8 & 25 \\
2 & 29.4 & 50.4 & +21.0 & 28 \\
3 & 29.4 & 48.0 & +18.6 & 26 \\
4 & 29.4 & 44.9 & +15.5 & 23\\
5 & 29.4 & 51.6 & +22.2 & 32 \\
\midrule
Mean / total & 29.4 & 48.2 & +18.8 & 134 \\
\bottomrule
\end{tabularx}
\par\smallskip
\begin{tabularx}{\linewidth}{@{}Yrrrrr@{}}
\toprule
\textbf{Inquiry group} & \shortstack{\textbf{Evaluated/}\\\textbf{attempts}} & \shortstack{\textbf{Mean}\\$\Delta$ (pp)} & \textbf{Up/Tie/Down} & \textbf{Reversals} & \shortstack{\textbf{Zero-progress}\\\textbf{loops}} \\
\midrule
Direct repair (F) & 51/86 & +8.7 & 25/6/20 & 10 & 8 \\
Untested hypothesis (H) & 17/29 & +11.2 & 11/2/4 & 2 & 0 \\
Diagnostic inquiry (I/G) & 11/15 & +14.3 & 5/3/3 & 1 & 1 \\
\bottomrule
\end{tabularx}
\end{table}

%% file: appendix_evidence.tex
\subsection{Additional Examples of Opposing Workflow Edits}\label{app:reversals}
These cases are drawn from one recorded Lean RSI trajectory and illustrate the patterns discussed in the introduction. They are not additional experimental groups. Version identifiers refer to this trajectory; the underlying quotations and selected task records are retained in \texttt{evidence/reversal\_cases.json} and \texttt{evidence/reversal\_witnesses.json}.

\paragraph{Failure history---more context or less anchoring?}
In $v_4$, a fresh-strategy final call solves row 30 after earlier attempts fail, although the aggregate ties its parent at 19/54. Later, $v_{13}$ supplies both rejected proofs and diagnostics to avoid repeating failed approaches. It ties its parent $v_{11}$ at 18/54 and is reverted; the agent attributes the lack of improvement to anchoring. In $v_{19}$, it instead removes rejected proofs and diagnostics entirely to increase independence. The score falls from parent $v_{18}$'s 20/54 to 17/54, and the change is reverted. The same difficulty---an unproductive final attempt---has prompted both more and less history. Neither trial separates harmful anchoring from useful diagnostic information, missing library knowledge, or response variation. There is no aggregate initial gain to claim for $v_4$; its benefit is one observed final-call conversion.

\paragraph{Tactic fallbacks---expansion without resolving applicability.}
In $v_3$, a checked \texttt{native\_decide} fallback directly accepts rows 4, 5, and 33; rows 5 and 33 were unsolved by its parent. The agent expands this branch in $v_{12}$ to include numeric and Aesop alternatives. No acceptance is attributed to the added branches, the aggregate remains 18/54, and the portfolio is reverted. Later, $v_{20}$ replaces the post-repair native check on quantified goals with \texttt{omega}. It directly solves no problem, its observed score is 18/54 versus parent $v_{18}$'s 20/54, and it is reverted. The newly accepted row 50 uses the third model call, not the static fallback. These trials demonstrate why a useful tool on one class of targets does not establish the applicability of a broader portfolio. They do not prove that trying an alternative tactic is intrinsically wasted effort.

%% file: appendix_protocol.tex
\subsection{Mechanism Record}
Each record contains the observed phenomenon, applicable task conditions, raw trace references, competing mechanism hypotheses, predicted outcomes, executed probe code and results, unresolved alternatives, a proposed workflow intervention, and a recurrence check. Hypotheses retain the status \emph{unresolved}, \emph{supported under tested conditions}, or \emph{contradicted}; no language-model confidence score alone changes a hypothesis to supported. An inconclusive probe and an unsuccessful implementation are distinct outcomes.

\subsection{Probe and Intervention Protocol}
Before executing a probe, the agent records the manipulated factor, held-fixed artifacts, observations expected under each live hypothesis, and its budget limit. Pure transformation replay is preferred when it can answer the question without another model call. When generation is necessary, the model snapshot, decoding settings, prompt, and random seeds are logged; repeated calls still require empirical variability checks. A positive result authorizes only the conclusion supported by the manipulation. Broader causal claims require additional controls.

Before implementing a workflow change, the agent specifies a mechanism-level prediction and a task-level acceptance criterion. For retrieval, the first may be retaining a needed lemma in context; the second is proof validity and task performance. An intervention can pass the first while failing the second because other proof errors remain. The record retains this distinction. New components expose their inputs, outputs, state, and tunable choices, but the framework does not prescribe the component type. Later edits are checked against the same task constraints and update the record when their behavior changes.

\subsection{Detailed Outcome Measurements}\label{app:method-measurements}
We report final held-out verification success, improvement over the shared seed, and success across recorded RSI checkpoints. Mechanism recurrence and repeated editing explain the trajectory of improvement. Errors are not grouped solely by their final message: a mechanism can survive a change of diagnostic. Unknown and multi-cause cases remain explicit, and independent reviewers audit intervention-supported labels without seeing the method identity.

For mechanism $m$, let $E_m$ be a fixed, independently specified set of eligible opportunities and $z_m(S,e)$ indicate its observed occurrence. The recurrence measurement is
\begin{equation}
R_m(S;E_m)=\frac{1}{|E_m|}\sum_{e\in E_m}z_m(S,e).
\end{equation}
Eligibility is fixed across the compared workflows; exposure coverage is reported separately when a changed workflow no longer reaches the relevant stage. Paired artifact replay isolates process effects, whereas complete reruns measure interactions with subsequent generation. Main-task pass rates report means and standard deviations across three outer runs; behavioral rates pool their underlying counts.

\subsection{Task Details and Main Evaluation Configuration}\label{app:task-settings}
\paragraph{Lean proof workflow.}
We retain the original RSI-Exam Lean proof dataset and supplement it with problems from VeriSoftBench~\citep{rsiexam2026tasks,xin2026verisoftbench}, yielding 92 training and 48 test problems in total. VeriSoftBench contributes software-verification proof obligations with project-specific definitions and cross-file dependencies. The initial proof workflow and optimization protocol follow RSI-Exam. Each task requires completing a fixed theorem in its associated context. The seed generates one candidate and permits one compiler-guided repair. We retain the per-problem limits of three model calls, 12,000 model tokens, four compiler checks, and 180 seconds, with a 24-hour outer optimization budget. Rejected constructs and budget overruns count as failures. Training problems provide feedback for workflow improvement; held-out outcomes are used only for evaluation. The Lean and library versions must match the selected VeriSoftBench project environments; RSI-Exam's original Lean 4.9 environment is not assumed to support these repositories.

\paragraph{Verus code generation.}
We select 113 training and 46 test problems carrying the \texttt{hard} label in VeriContest~\citep{xie2026vericontest}. The initial workflow follows the same generate--verify--repair structure, replacing Lean proof compilation with Verus verification. Both domains share the 24-hour outer budget and the RSI-Exam task-solving configuration, with domain-specific inputs and verifier interfaces.

\paragraph{Models and repeated runs.}
GPT-5.6-sol is the outer improvement agent. The inner model remains the RSI-Exam snapshot \texttt{gpt-5.4-2026-03-05}, with temperature 0 and seed 0. Each method starts from the same domain-specific $S_0$ and is evaluated over three independent outer runs. For final task pass rates, we compute the arithmetic mean and sample standard deviation across those runs. The 24-hour allowance includes workflow evaluation and, for \method{}, investigation and probe execution; final held-out evaluation is separate. The primary stopping budget is elapsed time; aggregate token cost is not reported.

Table~\ref{tab:main-manifest} summarizes the task and model configuration.
\begin{table}[htbp]
\centering\small
\caption{\textbf{Main experiment configuration.} The seed is the common RSI starting point. Both domains retain the RSI-Exam inner model and per-task limits.}
\label{tab:main-manifest}
\begin{tabularx}{\linewidth}{@{}YYY@{}}
\toprule
\textbf{Setting} & \textbf{Lean proving} & \textbf{Verus VCG} \\
\midrule
Benchmark / subset & RSI-Exam + VeriSoftBench & VeriContest, \texttt{hard} \\
Training / test problems & 92 / 48 & 113 / 46 \\
Outer improvement model & GPT-5.6-sol & GPT-5.6-sol \\
Inner model snapshot & \texttt{gpt-5.4-2026-03-05} & \texttt{gpt-5.4-2026-03-05} \\
Inner temperature / seed & 0 / 0 & 0 / 0 \\
Outer budget per run & 24 hours & 24 hours \\
Per-task model calls / tokens & 3 / 12,000 & 3 / 12,000 \\
Per-task checks / wall time & 4 / 180 seconds & 4 / 180 seconds \\
Independent outer runs & 3 & 3 \\
Reported aggregation & Mean $\pm$ sample SD & Mean $\pm$ sample SD \\
\bottomrule
\end{tabularx}
\end{table}

\subsection{Second-Order Behavior Analysis}\label{app:second-order-analysis}
\paragraph{Units and annotation.}
An investigation episode begins when the agent explicitly examines a reusable workflow mechanism behind an observed pattern, and includes subsequent inquiry into that same question until closure or the run ends. An episode can span several RSI rounds; a reopen is linked to its original record. Direct artifact fixes without mechanism inquiry are not episodes. We reconstruct episodes from the full traces of every method, including Naive RSI and the memory ablation, without relying on method-specific record IDs. Annotation follows Appendix~\ref{app:classification}; unknown cases remain explicit. Independent reviewers apply the same criteria to all variants and resolve disagreements with reference to recorded artifacts.

\paragraph{Tested inquiry and decision revision.}
The tested-inquiry rate is the fraction of all episodes, including unfinished ones, containing a mechanism explanation, an explicit testable prediction, and an executed diagnostic check capable of discriminating that explanation from an alternative. Ordinary post-edit pass-rate evaluation alone does not qualify. The decision-revision rate uses the same denominator and counts tested episodes with a traceable probe-result-to-decision link that changes an explanation, redirects an intervention, or withdraws a proposal. A changed prompt or parameter without such a link does not count. A high revision rate is not inherently desirable, since an initially correct explanation may survive testing.

\paragraph{G acceptance and gain.}
G edits are identified using the existing evidence-guided intervention definition, independently of whether the run invokes a module named G. Acceptance is accepted G edits divided by evaluated G edits. Gain is the arithmetic mean of development-pass-rate differences between each evaluated G candidate and its own parent, measured on the same problem set under matched task-solving limits; rejected candidates are included. Unevaluated edits are excluded and counted separately. This measures observed single-step gain, including interactions with inherited edits, rather than a causal allocation of the final trajectory improvement.

\paragraph{Mechanism recurrence.}
For each audited accepted G edit, freeze a mechanism-identification check and an eligible case set before comparing its parent and child workflows. Using the recurrence definition above, compute $100\,[R_m(S_{\mathrm{child}};E_m)-R_m(S_{\mathrm{parent}};E_m)]$ in percentage points. Average over audited interventions within each run. Report audit coverage, eligible case counts, and exposure changes, retaining unresolved or unobservable outcomes rather than treating them as resolved failures. The paired comparison measures immediate recurrence reduction. Audit results from held-out cases never feed back into RSI.

\paragraph{Aggregation.}
For the three behavioral rates, pool the reported numerator and denominator counts across runs and compute their ratio, separately for Lean and VCG. Table~\ref{tab:second-order-behavior} reports the supplied point estimates for the remaining metrics; run-level standard deviations are unavailable. A zero denominator is N/A rather than zero; if fewer than three runs are eligible, report the contributing count, and do not report an SD for a single observation. The underlying counts accompany rates in Appendix~\ref{app:behavior-counts}. Zero-progress loops retain the motivation-section definition.

\subsection{Reported Counts for Behavioral Rates}\label{app:behavior-counts}
Table~\ref{tab:behavior-counts} preserves the numerator/denominator counts underlying the percentages in Table~\ref{tab:second-order-behavior}. Each percentage is computed independently as $100n/d$ and rounded to one decimal place. Tested and Revised share the total investigation count as their denominator. Revised counts only decisions changed by executed diagnostic evidence, so Revised cannot exceed Tested. The ablation without executable probes disables both probe execution and probe-driven revision; these metrics are displayed as not applicable (---) in the tables.
\begin{table}[htbp]
\centering\small
\caption{\textbf{Underlying counts for behavioral rates.} Entries show numerator/denominator for each reported metric. Dashes indicate probe-dependent metrics that are not applicable without probes; the corresponding investigation totals are 2,017 (Lean) and 1,668 (VCG).}
\label{tab:behavior-counts}
\begin{tabularx}{\linewidth}{@{}Yrrr@{}}
\toprule
\textbf{Method} & \textbf{Tested} & \textbf{Revised} & \textbf{Accepted G} \\
\midrule
\multicolumn{4}{l}{\textit{Lean proving}} \\
Naive RSI & 273/1840 & 16/1840 & 34/62 \\
\method{} & 1482/3223 & 729/3223 & 51/76 \\
\quad w/o probes & --- & --- & 39/69 \\
\quad w/o memory & 1194/2870 & 438/2870 & 43/72 \\
\midrule
\multicolumn{4}{l}{\textit{Verus VCG}} \\
Naive RSI & 111/1249 & 42/1249 & 31/58 \\
\method{} & 1037/2491 & 514/2491 & 47/69 \\
\quad w/o probes & --- & --- & 35/62 \\
\quad w/o memory & 824/2186 & 326/2186 & 39/65 \\
\bottomrule
\end{tabularx}
\end{table}

%% file: appendix_implementation.tex
This appendix describes the controller, investigation records, and tool interfaces for the method in Section~\ref{sec:method}. Task-level settings are reported in Appendix~\ref{app:task-settings}.

\subsection{Controller and Model Calls}
A Python controller coordinates calls to one fixed base LLM with distinct contexts and instructions for detection, investigation, and editing. No additional training is required. The controller performs record retrieval, tool execution, persistence, and budget accounting. The outer and inner model configurations follow Table~\ref{tab:main-manifest}.

After an ordinary RSI attempt finishes, the controller logs its edit and outcome, checks for anomalies, and updates the corresponding investigation. It advances at most one investigation step before resuming ordinary RSI. A step forms or revises hypotheses, runs a probe, or hands a supported finding to G; it need not complete an investigation. Eligible investigations are processed by the oldest pending update to avoid repeatedly favoring the newest signal. A G proposal enters the next ordinary edit--evaluate cycle. Investigation memory starts empty, without a hand-authored library of mechanisms or fixes.

\subsection{Attempt Records and Anomaly Detection}
The compact record $X_t$ links to the following stored fields:
\begin{lstlisting}
attempt_id
parent_version, candidate_version
component, dimension, before, after
rationale, expected_outcome
  # expected_outcome is null if not explicitly stated
observed_outcome, evaluation_result, trace_paths
\end{lstlisting}
Code diffs and logs supply available structured fields; a short extraction call summarizes natural-language edits when necessary. Observations retain links to their source artifacts. Each modified dimension has an entry linked to the same attempt, preserving the distinction between a multi-part edit and independent trials.

The controller searches recent attempts for matching components or normalized failure types. If none are found, it skips historical semantic comparison, but still checks an explicit current prediction against the current outcome when both are available. Otherwise, it supplies the current record and a bounded set of related records to the detector prompt. The prompt requests an anomaly type, supporting attempt IDs, and the specific observations that justify the flag. Recurrence requires a targeted failure persisting or returning after a remedy; reversal requires opposing changes addressing a similar symptom; mismatch requires an explicit prediction contradicted by an observed outcome. Missing predictions or missing measurements cannot establish mismatch. Any detected signal enters memory without an additional root-cause investigation at this stage.

\subsection{Investigation Memory and Retrieval}
Each investigation is stored as a local JSON record with an ID, affected component, linked attempts, current question, hypotheses, completed probes and results, next check, and status. The statuses are \texttt{open}, \texttt{deferred}, \texttt{ready\_for\_G}, and \texttt{monitoring}. Hypothesis support is stored separately from investigation status.

For a new anomaly, the controller first retrieves investigations associated with the component. An LLM compares their questions with the new observation and chooses an existing ID or requests a new record. Matching compiler messages alone is insufficient to merge investigations. Memory writes append evidence references and preserve previous results, including negative results. A deferred record specifies what is already known and what remains to be checked; an accepted intervention moves the investigation to monitoring, while new conflicting observations can reopen it.

For hypothesis formation, historical attempts are filtered by component and ranked by embedding similarity between the current anomaly description and their edit rationales. The retrieved set includes both successes and failures when available; absent contrasts are not invented. The model receives these records, the investigation memory, and the implicated workflow code and prompts, with file-reading tools for additional inspection.

\subsection{Hypothesis Formation and External Knowledge}
The formation prompt asks for a workflow-level question and a small set of possible causes. Every hypothesis must supply:
\begin{lstlisting}
mechanism: workflow operation that may produce the failure
observations: supporting and conflicting trace references
prediction: observable consequence if the explanation holds
check: executable way to inspect that consequence
\end{lstlisting}
A proposed parameter change without a mechanism and a prediction does not complete this step. The hypotheses need not be mutually exclusive. For example, missing retrieval coverage and poor ranking can affect different cases within the same investigation. A validator checks required fields and source references; the diagnostic probe, rather than schema compliance, tests the substantive explanation.

External retrieval is optional. The model must identify a specific information need, such as the semantics of a retrieval API or a way to test whether a lemma applies to a goal. It searches project code and official documentation first, then research literature as needed. Queries, retrieved sources, and the hypothesis or check they inform are saved. External sources can suggest a cause or a test but cannot establish that the cause occurs in the current workflow. A model's assessment of lemma relevance likewise remains a hypothesis until checked.

\subsection{Probe Generation, Selection, and Execution}
Available tools read and compare files, execute Python analysis scripts, call Lean or Verus, and replay a selected workflow stage. For each proposed probe, the LLM specifies the hypothesis, input cases or artifacts, script or tool arguments, held-fixed conditions, predicted distinguishing observation, and execution cap. The controller checks paths and required fields and executes the probe in a temporary workspace or checkout, leaving the accepted workflow unchanged.

Probe selection prefers existing-log inspection, followed by verifier checks or stage replay, and requests new generations only when needed. Among affordable checks, the prompt asks for one whose possible outcomes would change the proposed intervention or distinguish unresolved explanations. This is a qualitative selection rule, not a learned estimate of information gain. A probe that cannot fit within the remaining research allowance is deferred.

The executor records stdout, stderr, exit status, timing, verifier outcomes, and any generated artifacts. A script failure is recorded as an execution problem, not evidence against the hypothesis. Directly measurable observations are extracted by code; an LLM compares them with the pre-recorded prediction and returns \emph{supported under tested conditions}, \emph{contradicted}, or \emph{inconclusive}, citing actual outputs. A confidence score alone cannot establish support.

At least one completed diagnostic check supporting an identifiable workflow intervention is required for handoff to G. This authorizes an edit proposal, not a claim that the root cause has been conclusively established. Contradictory or inconclusive results update the question and next check. When a meaningful comparison requires more repetitions than the remaining allowance, the result remains unresolved and the inquiry is deferred.

\subsection{Intervention Handoff}
G receives the hypothesis and its scope, executed probe code and results, implicated workflow files, expected behavioral change, and task constraints. The ordinary RSI editing agent generates a patch and its expected effect. Parameter changes, modifications to existing components, and new components are all permitted when justified by the diagnosis. The original development evaluator and acceptance rule decide whether to retain the patch; diagnostic support alone never merges it. Both accepted and rejected outcomes are appended to investigation memory. Held-out tasks do not guide these decisions.

%% file: appendix_ai_use.tex
Generative AI tools assisted with conceptual development, hypothesis refinement, experimental design, literature synthesis, interpretation of author-provided results, manuscript editing, and figure-generation code. The authors are responsible for the final text, claims, and results.

%% file: statements.tex
\section{Ethics Statement}\label{app:ethics}
This work studies automated agents on public formal-verification benchmarks. Generated proofs and program artifacts are evaluated with formal verifiers under fixed task constraints.

\section{Reproducibility Statement}\label{app:reproducibility}
Sections~\ref{sec:method}--\ref{sec:experiments} describe the method and experimental setup. The appendices provide implementation interfaces, task settings, annotation rules, and the counts underlying the reported behavioral rates.

%% file: related_context.tex
\section{Additional Related Work}
\label{app:related-context}
\paragraph{Reasoning supervision and information quality.}
IPS uses auxiliary questions to provide process supervision for short-video moderation~\citep{liu-etal-2026-ips}, while reasoning-enhanced domain-adaptive pretraining combines captioning, question answering, and reasoning traces~\citep{wang-etal-2025-reasoning-enhanced}.
Both improve task reasoning through model training; SO-RSI instead investigates the reusable workflow around a fixed model.
Latent-space broadening targets informative annotation candidates, and audio-enhanced vision-language modeling improves available multimodal evidence~\citep{sun2025audio}.
These are examples of changing information supply, a distinct intervention from rewriting instructions; our investigation seeks evidence for which kind of workflow change is needed.

\paragraph{Task-grounded evaluation.}
WorldCoder-Bench checks runtime states and transitions in generated 3D programs through executable probes~\citep{lu2026worldcoder}; SO-RSI uses probes to test explanations of workflow failures before constructing an intervention.
Outside code, ReVA tests scene-centric and temporal reasoning in remote-sensing videos rather than relying on repetitive static-image questions~\citep{yao2026revascenecentricdatasetrepetition}.
The survey by \citet{yao2026semantic} organizes language-guided segmentation from semantic matching through referring expressions to reasoning-based guidance.
These works distinguish the capabilities a task actually requires; our first-/second-order distinction instead describes whether improvement attempts investigate the process producing recurring failures.
RAGScope evaluates inexpensive evidence gates that route uncertain RAG outputs for further review~\citep{liu2026ragscope}.
Its routing objective relates to our lightweight detector, although our trigger concerns patterns across RSI attempts and initiates a persistent mechanism investigation.

\paragraph{Representing prior experience.}
In-context Time Series Predictor packages historical forecasting tasks as lookback--future examples for contextual prediction~\citep{lu2025ictsp}.
It illustrates an alternative to supplying an undifferentiated history: prior experience can be organized into task-relevant units.
SO-RSI uses a different unit---a mechanism hypothesis linked to its tests and outcomes---so that a later investigation can revisit what was supported or contradicted, rather than simply reuse an earlier response.

\paragraph{Mechanism-specific interventions in other domains.}
\citet{wu2026geocueformergeometryguidedwaveletrepresentation} address degraded underwater imagery with geometry-constrained frequency enhancement and prediction-cued refinement.
Their architecture is designed for a known source of interference; SO-RSI studies how an improvement agent can investigate a failure pattern before selecting a workflow intervention.

\paragraph{Levels of diagnosis and intervention.}
Model-internal analysis can recover interpretable activation directions using independent component analysis~\citep{liu2026ica}, while inference-time intervention can modify initial-token attention without updating model weights~\citep{han2026zerotuning}.
SO-RSI operates at a different level: its evidence comes from workflow attempts and diagnostic tests, and its interventions modify reusable control logic around the model.